\documentclass[]{fairmeta}
\usepackage[utf8]{inputenc}
\usepackage[T1]{fontenc}
\usepackage[round]{natbib}
\usepackage{hyperref}
\usepackage{url}
\usepackage{booktabs}
\usepackage{amsfonts}
\usepackage{amsmath}
\usepackage{amssymb}
\usepackage{amsthm}
\usepackage{nicefrac}
\usepackage{microtype}
\usepackage{xcolor}
\usepackage{algorithm}
\usepackage{algorithmic}
\usepackage{graphicx}
\usepackage{lmodern}
\graphicspath{{../experiments/figures/}}
\usepackage{pgfplots}
\pgfplotsset{compat=1.18}
\usepgfplotslibrary{groupplots}
\usepgfplotslibrary{fillbetween}
\usepackage{subcaption}
\usepackage{multirow}
\usepackage{wrapfig}

\usepackage{amsmath,amsfonts,bm}

\def\eqref#1{equation~\ref{#1}}

\def\1{\bm{1}}

\DeclareMathAlphabet{\mathsfit}{\encodingdefault}{\sfdefault}{m}{sl}
\SetMathAlphabet{\mathsfit}{bold}{\encodingdefault}{\sfdefault}{bx}{n}

\title{When LLM Agents Fail to Read the Room: ReAdapt for Relational Social Reasoning}

\author[]{Jianzhe Lin, Xiaolin Li, Yunda Liu, Fei Wang, Jubin Chheda}

\affiliation[1]{MetaAI}
\contribution[*]{Work done at Meta}

\abstract{A social agent's most fundamental decisions (for example, ``should I react to this
post?'', ``who should I reach out to?'') are not purely content problems.
The appropriate action can hinge on the \emph{latent relationship} between
people, i.e., tie strength, reciprocity, and mutual connections, rather than on
which piece of content is most salient. Yet standard LLM-based agent loops do
not explicitly represent how newly observed relational evidence should revise
the agent's current social hypothesis, making them vulnerable to
surface-obvious choices when relational and content cues diverge.

We formalize this failure mode through a \textbf{relationship-reasoning
benchmark}: 500 synthetic social worlds containing friendships, directed
follows, reaction histories, and feeds, yielding 1,000 queries across two
tasks---\emph{reaction selection} (``which post should I engage with?'') and
\emph{warm introduction} (``who is the best bridge to this person?'').
By construction, the surface-obvious candidate differs from the
relationship-grounded oracle in approximately $53\%$ of queries, forming an
\emph{overturn} subset in which the agent must use relational evidence to
revise an initially plausible choice.

We propose \textbf{ReAdapt}
(\textbf{Re}lationship-\textbf{Ad}aptive \textbf{A}gent with
\textbf{P}olicy-driven s\textbf{T}ate), which augments the ReAct loop with an
explicit structured social state
$z=(\mathcal{G},\mathcal{B},\mathcal{R},\mathcal{N},\mathcal{D})$
capturing goal, belief, relationship, norm, and disclosure. After each tool
observation, ReAdapt performs a typed \emph{Adapt} step that updates this state
and emits a policy operation
(\textsc{continue}, \textsc{switch}, \textsc{abandon}, or
\textsc{clarify}) before selecting the next action.

Evaluated with Gemini-3-Flash on a stratified subset of $n=150$ queries per
task, ReAdapt improves warm-introduction accuracy from $37\%$ to $51\%$
(\textbf{+14 points}) and reaction-selection accuracy from $69\%$ to $77\%$
(\textbf{+8 points}). Oracle regret decreases from $0.260$ to $0.152$ on
warm introduction and from $0.095$ to $0.053$ on reaction selection. With the
same underlying model, tool interface, and social environments, these results
suggest that explicit relational-state adaptation helps LLM agents translate
retrieved social evidence into revised decisions.}

\date{\today}
\correspondence{Jianzhe Lin at \email{jianzhelin@meta.com}}

\begin{document}

\maketitle

\section{Introduction}
\label{sec:introduction}
\begin{figure}[t]
    \centering
    \includegraphics[width=\linewidth]{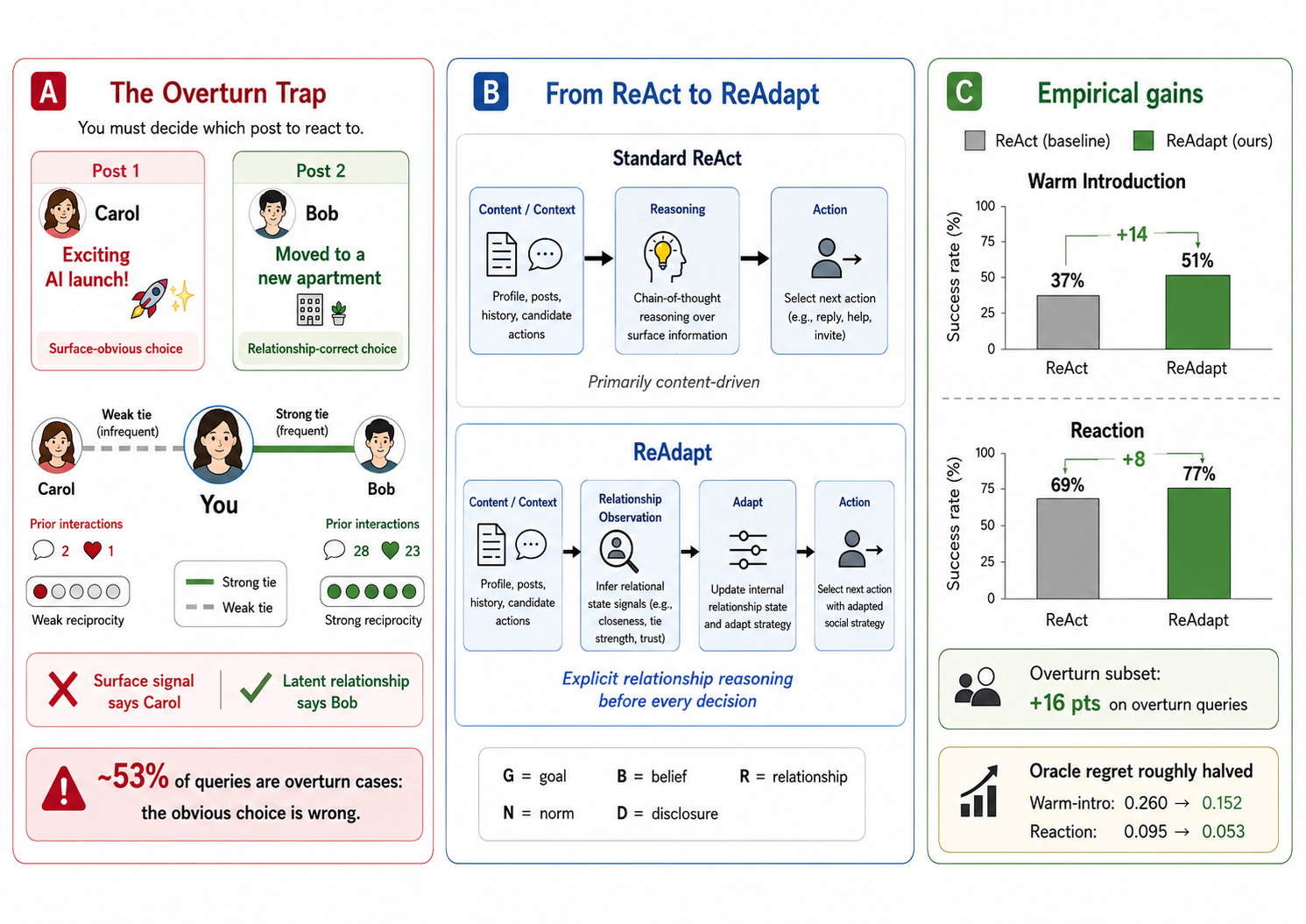}
    \caption{
    Explicit Relationship Reasoning Enables ReAdapt to Overturn Misleading Surface Signals.
    }
    \label{fig:motivation}
\end{figure}

A generally capable agent acting on behalf of a person must eventually solve
problems that are fundamentally social: Who should I reconnect with? Whom
should I ask for an introduction? Should I acknowledge this message now or
leave it alone? Which relationship deserves attention?

These questions are deceptively difficult. Their answers are rarely determined
by the literal content of the current message alone. Instead, they depend on
what might colloquially be called \emph{social know-how}: the largely implicit
knowledge of relationship history, reciprocity, obligations, norms, boundaries,
and timing that makes the same action appropriate toward one person and
inappropriate toward another. The difficult part of social intelligence is
often not understanding \emph{what was said}, but understanding
\emph{what the relationship makes appropriate}.

Existing work has demonstrated substantial progress in tool-using agents,
long-horizon reasoning, memory, and social simulation
\citep{yao2023react,schick2023toolformer,shinn2023reflexion,
wang2023voyager,park2023generative,zhou2023sotopia}. Other work evaluates
social commonsense, personas, and Theory of Mind: whether language models can
infer intentions, beliefs, or likely reactions
\citep{sap2019socialiqa,kosinski2023theory,ullman2023large,kim2023fantom}.
These capabilities are important, but relational judgment poses an additional
problem. An agent may correctly understand a post, correctly retrieve who
knows whom, and even correctly describe the strength of a relationship, yet
still fail to let this information change what it ultimately does.

Consider a simple example. A user's feed contains an exciting post about a
favorite topic from a casual acquaintance and an ordinary life update from a
close friend with whom the user has recently lost touch. The former is
immediately salient from content alone. The latter may nevertheless be the
more appropriate relationship-grounded choice: responding could reciprocate
past engagement or reinforce a valuable but dormant tie. Behavioral history,
reciprocity, and network structure have long been associated with
interpersonal tie strength in sociology and computational social science
\citep{granovetter1973strength,gilbert2009predicting}, yet these signals are
typically distributed across multiple pieces of evidence rather than stated
explicitly in the current content.

This setting exposes a subtle limitation of standard LLM-agent loops.
ReAct~\citep{yao2023react}, for example, interleaves reasoning, actions, and
environment observations. Importantly, we do \emph{not} claim that ReAct
fails to reason over observations. An observation is simply externally
returned evidence---for example, a tool reporting that Alice has reacted to
Bob's posts 17 times. The model may reason extensively about that observation.
The distinction we study is instead whether the effect of that evidence on
the agent's \emph{current hypothesis} is represented explicitly.

Formally, a standard ReAct-style policy may choose the next action $a_{t+1}$ directly
from the accumulated interaction history:
\begin{equation}
a_{t+1}
=
\pi_{\theta}
\left(
x,
h_t,
o_t
\right),
\label{eq:intro-react}
\end{equation}
where $x$ is the request, $h_t$ is the previous interaction history, and
$o_t$ is the latest environment observation. The observation is available to
the model, but the architecture does not require an explicit transition from
a previous social hypothesis to a revised one. Its influence is mediated
implicitly through the language-model context.

This distinction matters because retrieving evidence and revising a decision
are not the same operation. An agent may initially prefer candidate $A$,
observe evidence strongly favoring candidate $B$, mention that evidence in its
reasoning, and nevertheless retain $A$ as its final choice. We refer to this
as a \textbf{failure of evidence-to-decision coupling}: relevant information
has been acquired, but has not reliably altered the policy.

We study this phenomenon through a controlled
\textbf{relationship-reasoning benchmark}. We generate 500 synthetic social
worlds containing profiles, friendships, directed follows, reaction histories,
group memberships, and posts, yielding 1,000 queries across two tasks:
\emph{reaction selection} and \emph{warm introduction}. Rather than attempting
to define universally correct human social behavior, the benchmark specifies
a transparent \emph{relationship-grounded oracle} based on interpretable
relational features.

For every query, we also identify a \emph{surface-obvious} candidate: the
choice favored by immediately salient content or local graph structure. We
define an \emph{overturn case} when this candidate differs from the
relationship-grounded oracle:
\begin{equation}
\mathrm{Overturn}(q)
=
\mathbb{I}
\left[
a_q^{\mathrm{surf}}
\neq
a_q^{*}
\right].
\end{equation}
Approximately $53\%$ of the full benchmark satisfies this condition. These
examples isolate the behavior of interest: success cannot come from simply
following the initial surface heuristic; the agent must acquire additional
relational evidence and allow that evidence to revise its choice.

We introduce \textbf{ReAdapt}
(\textbf{Re}lationship-\textbf{Ad}aptive \textbf{A}gent with
\textbf{P}olicy-driven s\textbf{T}ate) to make this revision explicit.
ReAdapt maintains a structured social state
\begin{equation}
z_t
=
\left(
\mathcal{G}_t,
\mathcal{B}_t,
\mathcal{R}_t,
\mathcal{N}_t,
\mathcal{D}_t
\right),
\label{eq:intro-state}
\end{equation}
capturing the agent's current representation of goal, belief, relationship,
norm, and disclosure.

After receiving observation $o_t$, ReAdapt requires an explicit state
transition:
\begin{equation}
z_{t+1}
=
\operatorname{Adapt}_{\theta}
\left(
z_t,
o_t,
x
\right).
\label{eq:intro-adapt}
\end{equation}
The Adapt step records not merely a new interpretation, but the
\emph{change} induced by the observation. It then emits one of four typed
policy operations---\textsc{continue}, \textsc{switch},
\textsc{abandon}, or \textsc{clarify}---which conditions the next action:
\begin{equation}
a_{t+1}
=
\pi_{\theta}
\left(
x,
h_{t+1},
z_{t+1},
\omega_{t+1}
\right).
\end{equation}

Thus, the contrast between ReAct and ReAdapt is not
\emph{observation versus reasoning}, nor do we claim that ReAct cannot reason
about relationships. Both systems receive and reason over the same external
evidence. The distinction is between
\emph{implicit evidence integration} and an
\emph{explicit, policy-relevant state transition}. Put differently,
\textbf{ReAdapt does not change what the agent observes; it changes how
observations become state}.

This formulation connects to broader work on structured reasoning and agent
memory. Chain-of-Thought, self-consistency, Tree of Thoughts, and search-based
agent methods expose or explore intermediate reasoning
\citep{wei2022cot,wang2023selfconsistency,yao2023tot,zhou2023lats}, while
Reflexion and memory-oriented architectures introduce mechanisms for storing
and reusing information across reasoning steps or episodes
\citep{shinn2023reflexion,sumers2023coala,packer2023memgpt,
zhong2024memorybank}. ReAdapt addresses a complementary question:
\emph{when new evidence contradicts the agent's current social hypothesis,
does the architecture provide an explicit mechanism for changing that
hypothesis before acting?}

We evaluate ReAct and ReAdapt with Gemini-3-Flash on a stratified
300-query subset of the benchmark, with 150 queries from each task. Both
conditions receive the same model, underlying social worlds, candidates,
tool interface, and maximum tool-call budget. ReAdapt improves
warm-introduction accuracy from $37\%$ to $51\%$ and reaction-selection
accuracy from $69\%$ to $77\%$. The largest observed improvement occurs on
warm-introduction overturn cases, where accuracy rises from $33\%$ to $49\%$.

We additionally report \textbf{benchmark-oracle regret}, which captures the
severity of an error rather than only whether the top-ranked candidate was
selected:
\begin{equation}
\mathrm{Regret}(q)
=
S_q(a_q^{*})
-
S_q(\hat{a}_q).
\label{eq:intro-regret}
\end{equation}
A wrong choice that is nearly as highly valued as the oracle optimum therefore
incurs low regret, whereas a substantially inferior choice incurs high regret.
ReAdapt reduces mean regret from $0.260$ to $0.152$ on warm introduction and
from $0.095$ to $0.053$ on reaction selection. We emphasize that this is
regret with respect to the benchmark oracle, not an estimate of universal
human social utility.

Our contributions are threefold:
\begin{enumerate}
    \item We identify \textbf{evidence-to-decision coupling}---the ability to
    let newly acquired relational evidence revise an existing choice---as a
    central challenge for socially adaptive LLM agents, and introduce a
    benchmark of 500 social worlds and 1,000 queries with a controlled
    \emph{overturn} diagnostic.

    \item We introduce \textbf{ReAdapt}, a lightweight extension to ReAct that
    converts observations into explicit changes in a structured social state
    before subsequent action.

    \item In a controlled 300-query proof-of-concept evaluation, we show that
    ReAdapt improves relationship-grounded decisions on two social tasks,
    including a 16-point gain on the most challenging warm-introduction
    overturn cases and substantial reductions in benchmark-oracle regret.
\end{enumerate}

\section{Relationship-Reasoning Benchmark}
\label{sec:benchmark}

\subsection{Synthetic Social Worlds}

We generate \textbf{500 synthetic social worlds}. Each world is a
fully specified social environment containing 8--15 users and one designated
\emph{ego}, on whose behalf the agent acts. A world contains:

\begin{itemize}
    \item \textbf{Profiles}: names, short descriptions, and interest tags;
    \item \textbf{Friendships}: undirected social ties;
    \item \textbf{Follows}: directed attention relationships;
    \item \textbf{Reaction histories}: directed interaction counts between
    pairs of users;
    \item \textbf{Group memberships}: shared community affiliations; and
    \item \textbf{Feeds}: authored posts with topics and text.
\end{itemize}

Worlds are generated procedurally to vary graph structure, density,
reciprocity, interaction history, and the alignment between surface-level and
relational evidence. The benchmark generator has access to the complete
environment, whereas the evaluated agent must acquire relevant evidence
through tools.

\subsection{Tasks}

The benchmark contains \textbf{1,000 queries}, with 500 queries for each task.

\paragraph{Reaction selection.}
The ego observes several candidate posts and must decide which post to engage
with. The benchmark assigns candidate post $p$, authored by user $a_p$, the
score

\begin{equation}
\begin{split}
S_{\mathrm{react}}(u,p)
={}&
0.28 F(u,a_p)
+0.24 M(u,a_p)
+0.15 H(u,a_p) \\
&+0.08 R(u,a_p)
+0.06 C(u,p),
\end{split}
\label{eq:reaction-oracle}
\end{equation}

where $F$ represents friendship, $M$ mutual-follow evidence, $H$ prior
reaction history, $R$ reciprocity, and $C$ content affinity.

The coefficients in Eq.~\eqref{eq:reaction-oracle} are benchmark design
parameters rather than an estimate of a universal human social utility
function. Relational signals such as interaction history, reciprocity, and
network structure are motivated by prior computational work on interpersonal
tie strength~\citep{granovetter1973strength,gilbert2009predicting}. We use
them here to construct a diagnostic setting in which relational evidence can
conflict with an immediately salient content cue.

\paragraph{Warm introduction.}
The ego wishes to connect with a target user and must select an intermediary
from several candidate bridges. Candidate quality depends on the relational
evidence along the ego--bridge--target path, including the ego--bridge
relationship, the bridge--target relationship, and shared social context.
Unlike reaction selection, warm introduction therefore requires reasoning
over indirect relational structure rather than only a direct ego--author tie.

\subsection{The Overturn Diagnostic}
\label{sec:overturn}

For each query $q$, let $a_q^{*}$ denote the highest-scoring candidate under
the relationship-grounded benchmark oracle. We separately define a
deterministic surface heuristic that selects an initially salient candidate
$a_q^{\mathrm{surf}}$. For reaction selection, the heuristic prioritizes
surface-level content salience; for warm introduction, it prioritizes an
immediately obvious bridge.

We define an overturn case as

\begin{equation}
\mathrm{Overturn}(q)
=
\mathbb{I}
\left[
a_q^{\mathrm{surf}}
\neq
a_q^{*}
\right].
\label{eq:overturn}
\end{equation}

Approximately $53\%$ of the complete benchmark consists of overturn cases.
These queries are particularly diagnostic because following the surface
heuristic is insufficient under the benchmark objective. Successful behavior
requires the agent to gather additional relational evidence and revise an
initially plausible choice.

The remaining queries are \emph{straight} cases, in which
$a_q^{\mathrm{surf}}=a_q^{*}$. Reporting performance separately on overturn
and straight subsets helps distinguish general task competence from the
specific ability to revise a misleading initial hypothesis.

\subsection{Tool Interface}

The agent investigates each social world through seven tools:

\begin{itemize}
    \item \texttt{view\_profile(id)}: retrieve a user's profile and interests;
    \item \texttt{list\_friends(id)}: retrieve a user's friend set;
    \item \texttt{are\_friends(a,b)}: query whether two users are friends;
    \item \texttt{mutual\_friends(a,b)}: retrieve shared friends;
    \item \texttt{follow\_status(a,b)}: retrieve directed and mutual-follow
    status;
    \item \texttt{reaction\_history(actor,target)}: retrieve prior interaction
    counts; and
    \item \texttt{view\_post(post\_id)}: retrieve post metadata and content.
\end{itemize}

Each episode permits at most 24 tool calls. Both ReAct and ReAdapt receive the
same tool interface and access to the same underlying social environment.
Thus, the relevant information is available to both agents; the challenge is
deciding what evidence to retrieve and how to use it in subsequent decisions.

\subsection{Evaluation Metrics}

We report three primary metrics.

\paragraph{Pick accuracy.}
For predicted candidate $\hat{a}_q$,

\begin{equation}
\mathrm{Accuracy}
=
\frac{1}{N}
\sum_{q=1}^{N}
\mathbb{I}
\left[
\hat{a}_q=a_q^{*}
\right].
\end{equation}

\paragraph{Overturn accuracy.}
We separately report accuracy on the subset satisfying
$\mathrm{Overturn}(q)=1$. This directly measures whether an agent can recover
when the surface-level candidate conflicts with the benchmark oracle.

\paragraph{Oracle regret.}
Accuracy treats all incorrect candidates equally, so we additionally measure

\begin{equation}
\mathrm{Regret}(q)
=
S_q(a_q^{*})
-
S_q(\hat{a}_q),
\label{eq:regret}
\end{equation}

where lower values are better. Regret captures whether an incorrect prediction
is nevertheless close to the oracle optimum.

\section{Method: ReAdapt}
\label{sec:method}

ReAdapt augments the standard observation--action loop with an explicit
\textbf{Adapt} stage. Its central design principle is that a newly retrieved
observation should not merely become another piece of text in the interaction
history; it should explicitly update the agent's current model of the social
situation before the next decision is made.

\subsection{Baseline: ReAct}

ReAct~\citep{yao2023react} interleaves language-model reasoning with
environment actions. Let $x$ denote the user request and let
$h_t=(a_{<t},o_{<t})$ denote the interaction history before step $t$.
A ReAct agent selects

\begin{equation}
a_t
=
\pi_{\theta}
\left(
x,
h_t
\right),
\label{eq:react-policy}
\end{equation}

and receives an observation

\begin{equation}
o_t
=
\mathrm{Env}(a_t).
\end{equation}

In our setting, this allows the model to inspect profiles, friendships,
follows, reaction histories, and posts before producing a final candidate.
However, the interaction loop imposes no explicit representation of how each
observation changes the agent's current relational hypothesis.

This distinction is important because decisive social evidence is often
distributed across multiple observations. An agent may learn friendship from
one tool call, reciprocity from another, and interaction history from a third.
Standard ReAct makes these observations available in context, but leaves their
integration and any resulting hypothesis revision to free-form model
reasoning.

\subsection{Structured Social State}

ReAdapt maintains an explicit state

\begin{equation}
z_t
=
\left(
\mathcal{G}_t,
\mathcal{B}_t,
\mathcal{R}_t,
\mathcal{N}_t,
\mathcal{D}_t
\right),
\label{eq:social-state}
\end{equation}

with five typed dimensions:

\begin{itemize}
    \item $\mathcal{G}$ (\textbf{Goal}): the inferred objective behind the
    user's request;
    \item $\mathcal{B}$ (\textbf{Belief}): known social facts and unresolved
    uncertainty;
    \item $\mathcal{R}$ (\textbf{Relationship}): tie strength, reciprocity,
    and interaction evidence;
    \item $\mathcal{N}$ (\textbf{Norm}): relevant social-appropriateness
    constraints; and
    \item $\mathcal{D}$ (\textbf{Disclosure}): constraints on what information
    may appropriately be revealed.
\end{itemize}

The present benchmark primarily exercises the belief and relationship
dimensions. Goal, norm, and disclosure provide an interface for richer social
settings, but we do not claim that the current experiments independently
validate the necessity of all five dimensions.

\subsection{Adapt Before Acting}

After receiving observation $o_t$, ReAdapt performs

\begin{equation}
z_{t+1}
=
\operatorname{Adapt}_{\theta}
\left(
z_t,
o_t,
x
\right).
\label{eq:adapt-update}
\end{equation}

The Adapt output is structured rather than unrestricted reflection. It records
a concise assessment of the relevant state dimensions, the \emph{delta}
relative to the previous state, and a policy operation

\begin{equation}
\omega_{t+1}
\in
\left\{
\text{\textsc{continue}},
\text{\textsc{switch}},
\text{\textsc{abandon}},
\text{\textsc{clarify}}
\right\}.
\label{eq:operations}
\end{equation}

The operations have the following semantics:

\begin{table}[t]
\centering
\small
\caption{Policy operations emitted by the Adapt step.}
\label{tab:operations}
\begin{tabular}{@{}ll@{}}
\toprule
\textbf{Operation} & \textbf{Semantics} \\
\midrule
\textsc{continue}
& Retain the current hypothesis and gather more evidence \\
\textsc{switch}
& Revise the preferred candidate after contradictory evidence \\
\textsc{abandon}
& Do not execute the contemplated social action \\
\textsc{clarify}
& Request additional information before deciding \\
\bottomrule
\end{tabular}
\end{table}

The next action is conditioned on the updated state and operation:

\begin{equation}
a_{t+1}
=
\pi_{\theta}
\left(
x,
h_{t+1},
z_{t+1},
\omega_{t+1}
\right).
\label{eq:readapt-policy}
\end{equation}

The central distinction is therefore not that ReAdapt receives additional
relational information. Both agents have access to the same tools. Rather,
ReAdapt requires newly retrieved evidence to explicitly modify a persistent
representation before subsequent action.

\subsection{From Evidence Retrieval to Hypothesis Revision}

The \textsc{switch} operation is particularly relevant to overturn cases.
Suppose an agent initially favors candidate $A$ because $A$ is highly salient
under the surface heuristic. Subsequent tool calls reveal substantially
stronger relational evidence for candidate $B$. A standard ReAct agent must
translate these observations into a changed choice through unconstrained
reasoning. ReAdapt instead requires the new evidence to produce an explicit
state delta and, when warranted, a \textsc{switch} operation.

This motivates our central hypothesis:

\begin{quote}
\emph{Explicit relational-state adaptation should be especially useful when
new evidence must revise an initially plausible surface-level choice.}
\end{quote}

The benchmark's overturn subset provides a direct diagnostic for this
behavior.

\subsection{Implementation}

Both ReAct and ReAdapt are implemented as prompted, single-episode agents
without fine-tuning. Gemini-3-Flash is used in both conditions. The agents
receive the same query, candidate set, social world, tool interface, and
maximum budget of 24 tool calls.

The experimental intervention is the reasoning protocol. ReAct follows a
standard tool-using agent loop, whereas ReAdapt is instructed to maintain the
state in Eq.~\eqref{eq:social-state}, emit an Adapt update after each
observation, and select a policy operation from Eq.~\eqref{eq:operations}.

ReAdapt generates additional structured reasoning tokens, so the two
conditions are matched in model and information access but not in total
token-level computation.

\section{Experiments}
\label{sec:experiments}

\subsection{Experimental Setup}

\paragraph{Evaluation subset.}
The complete benchmark contains 1,000 queries. We perform frontier-model agent
evaluation on a \textbf{stratified subset of 300 queries}: 150 reaction
queries and 150 warm-introduction queries. The subset is constructed to
include both overturn and straight examples, enabling direct comparison of
the two reasoning regimes under both conditions.

Accordingly, our experiment should be interpreted as a controlled
proof-of-concept comparison rather than an exhaustive measurement over all
1,000 benchmark queries.

\paragraph{Model.}
Both conditions use Gemini-3-Flash with temperature $0$.

\paragraph{Controls.}
ReAct and ReAdapt are evaluated on identical queries and receive the same
model, social worlds, candidate sets, tool definitions, and maximum budget of
24 tool calls. The manipulated variable is the structured Adapt reasoning
protocol.

\subsection{Main Results}

Table~\ref{tab:main} presents the primary comparison.

\begin{table*}[t]
\centering
\small
\caption{
Performance on the stratified 300-query evaluation subset
($n=150$ per task). \textbf{Overturn} denotes queries for which the
surface-obvious candidate differs from the relationship-grounded oracle.
Higher accuracy and lower regret are better.
}
\label{tab:main}
\begin{tabular}{@{}llcccc@{}}
\toprule
\textbf{Task}
& \textbf{Agent}
& \textbf{Overall Acc.}
& \textbf{Overturn Acc.}
& \textbf{Straight Acc.}
& \textbf{Oracle Regret} \\
\midrule
Warm Introduction
& ReAct
& 37\%
& 33\%
& 42\%
& 0.260 \\
Warm Introduction
& \textbf{ReAdapt}
& \textbf{51\%}
& \textbf{49\%}
& \textbf{54\%}
& \textbf{0.152} \\
\midrule
Reaction
& ReAct
& 69\%
& 70\%
& 68\%
& 0.095 \\
Reaction
& \textbf{ReAdapt}
& \textbf{77\%}
& \textbf{76\%}
& \textbf{78\%}
& \textbf{0.053} \\
\bottomrule
\end{tabular}
\end{table*}

ReAdapt improves overall performance on both tasks. Warm-introduction accuracy
increases from $37\%$ to $51\%$, an absolute gain of 14 percentage points.
Reaction-selection accuracy increases from $69\%$ to $77\%$, an 8-point gain.

The largest observed improvement occurs on warm-introduction overturn cases.
ReAdapt improves from $33\%$ to $49\%$, a 16-point absolute gain. Reaction
selection shows a smaller but directionally consistent 6-point improvement on
overturn cases, from $70\%$ to $76\%$. These results are consistent with the
hypothesis that explicit state adaptation can help when an agent must revise a
plausible surface-level choice.

ReAdapt also improves straight-case accuracy: from $42\%$ to $54\%$ on warm
introduction and from $68\%$ to $78\%$ on reaction selection. The benefit is
therefore not restricted to explicit reversals, suggesting that structured
state may also support evidence integration when surface and relational cues
already agree.

Oracle regret provides a complementary view. On warm introduction, regret
decreases from $0.260$ to $0.152$, corresponding to a $41.5\%$ reduction. On
reaction selection, regret decreases from $0.095$ to $0.053$, a $44.2\%$
reduction. Thus, even when ReAdapt does not recover the highest-scoring
candidate, its selected candidate tends to lie closer to the benchmark
optimum.

\subsection{Task Difficulty}

Warm introduction is substantially harder than reaction selection for both
agents. The ReAct baseline achieves $37\%$ accuracy on warm introduction
compared with $69\%$ on reaction selection.

One plausible explanation is relational depth. Reaction selection primarily
requires evaluating direct ego--author evidence, whereas warm introduction
requires integrating both ego--bridge and bridge--target relationships. The
present experiment is consistent with this explanation, although it does not
causally isolate graph depth as the source of the difficulty.

\section{Analysis}
\label{sec:analysis}

\subsection{From Retrieved Evidence to Revised Decisions}

Qualitative inspection of successful ReAdapt trajectories on overturn cases
reveals the behavior the method is designed to support. A typical trajectory
has the following form:

\begin{enumerate}
    \item a surface cue makes candidate $A$ initially plausible;
    \item tool calls reveal stronger relational evidence for candidate $B$;
    \item the Adapt step updates the corresponding relational state;
    \item the policy operation changes to \textsc{switch}; and
    \item the agent investigates or confirms $B$ before finalizing.
\end{enumerate}

The important distinction is not simply whether relational evidence is
retrieved. In inspected ReAct failures, the model can mention evidence
favoring an alternative candidate while still retaining its initial choice.
ReAdapt externalizes the revision step by requiring the agent to state what
changed in its social state and whether that change warrants a policy update.

We treat this trajectory analysis as illustrative rather than as a
population-level causal decomposition. The current experiment evaluates the
complete ReAdapt protocol and does not isolate the contribution of individual
state dimensions or policy operations.

\subsection{Tool Usage}

ReAct uses 8.2 tool calls per episode on average, compared with 9.4 for
ReAdapt. Both remain well below the shared maximum of 24 calls.

The difference indicates that the two systems are not perfectly compute
matched. ReAdapt may gather modestly more evidence in addition to structuring
that evidence differently. Nevertheless, both agents have the same maximum
information-access budget and identical available tools. We therefore
interpret the experiment as a comparison of reasoning protocols under matched
model and environment access, rather than as a claim of equal computational
cost.

\section{Conclusion}
\label{sec:conclusion}

We introduced ReAdapt, a relationship-adaptive agent framework that augments
the ReAct loop with explicit structured social-state updates and typed policy
operations. We also introduced a controlled relationship-reasoning benchmark
containing 500 synthetic social worlds and 1,000 queries across reaction
selection and warm introduction.

In a stratified 300-query evaluation using Gemini-3-Flash, ReAdapt improves
accuracy from $37\%$ to $51\%$ on warm introduction and from $69\%$ to $77\%$
on reaction selection. Its largest observed gain is 16 percentage points on
warm-introduction overturn cases, where the surface-obvious candidate
conflicts with the relationship-grounded oracle. Oracle regret also decreases
substantially on both tasks.

These results support a simple design principle: access to the right social
evidence is not sufficient if that evidence does not reliably revise the
agent's current decision. Socially relevant observations should not merely
accumulate in context---they should \emph{change state}. ReAdapt provides one
lightweight mechanism for making that transition explicit.

\bibliographystyle{unsrtnat}
\bibliography{references}

\appendix

\section{Benchmark Sanity Checks}

We validate the evaluation harness using two scripted responders:

\begin{itemize}
    \item \textbf{Oracle responder}: always selects
    \texttt{answer}, producing $100\%$ pick accuracy.
    \item \textbf{Surface responder}: always selects
    \texttt{first\_obvious}, producing $100\%$ accuracy on straight cases and
    $0\%$ accuracy on overturn cases.
\end{itemize}

These checks confirm that the evaluation code implements the intended
relationship between the oracle answer, surface candidate, and overturn flag.
They are software-level sanity checks rather than evidence that the oracle
captures all dimensions of real-world social preference.

\section{Related Work}
\label{sec:related}

\paragraph{LLM-based social agents.}
LLM agents have increasingly been studied in simulated social environments.
Generative Agents~\citep{park2023generative} demonstrated that agents equipped
with memory, reflection, and planning can produce coherent long-horizon social
behavior. Subsequent work has explored larger social simulations, personas,
cooperation, negotiation, and multi-agent interaction.

Our setting differs in its unit of analysis. We do not ask whether a
population of agents produces realistic emergent social behavior. Instead, we
study whether a single agent acting on behalf of a user can make a
relationship-grounded decision when surface and relational evidence conflict.

\paragraph{Reasoning and state in language-model agents.}
ReAct~\citep{yao2023react} established an influential paradigm for
interleaving language-model reasoning and tool use. Reflexion
\citep{shinn2023reflexion} adds verbal reflection over previous attempts,
while Chain-of-Thought and self-consistency
\citep{wang2023selfconsistency} improve reasoning through alternative
inference procedures. More general cognitive-agent frameworks have also
emphasized explicit memory and state representations.

ReAdapt is complementary to these approaches. It does not introduce a new
general-purpose search or memory algorithm. Instead, it specializes the agent
state to relational decision making and requires observations to produce
explicit state changes before subsequent action.

\paragraph{Theory of Mind and social cognition.}
A growing literature evaluates whether LLMs can reason about beliefs,
intentions, and other latent social variables
\citep{kosinski2023theory,ullman2023large}. ReAdapt asks a
related but different question. Rather than testing whether a model can answer
a social-reasoning question in isolation, we ask whether an agent can
\emph{use} social evidence accumulated across tool calls to revise an action.

\paragraph{Computational relationship modeling.}
Computational social science has long modeled interpersonal ties using
interaction frequency, reciprocity, and graph structure.
Granovetter~\citep{granovetter1973strength} established the importance of tie
structure for social behavior, and later work showed that behavioral and
network features can predict interpersonal tie strength in online social
networks~\citep{gilbert2009predicting}.

Our contribution is not a new estimator of human tie strength. We instead turn
relational evidence into an active agent problem: the model must discover the
relevant evidence through tools and allow that evidence to influence a
subsequent decision.

\section{Limitations and Scope}
\label{sec:limitations}

\paragraph{Synthetic social worlds.}
The benchmark uses procedurally generated social environments rather than real
user relationships. This provides precise control over latent structure and
avoids privacy concerns, but necessarily simplifies the ambiguity and
idiosyncrasy of real interpersonal decisions.

\paragraph{Designed oracle.}
The benchmark oracle operationalizes a relationship-grounded objective chosen
for this study. We do not claim that its particular weighting represents a
universal definition of socially optimal behavior. The benchmark instead asks
a narrower question: whether an agent can recover a specified relational
objective when surface evidence points elsewhere.

\paragraph{Partial benchmark evaluation.}
Although the benchmark contains 1,000 queries, our frontier-model evaluation
uses a stratified subset of 300 queries. The reported results should therefore
be interpreted as a controlled proof-of-concept comparison rather than a
complete characterization of performance over the full benchmark.

\paragraph{Single model.}
We evaluate Gemini-3-Flash only. The magnitude of the effect may vary across
model families, scales, and inference systems.

\paragraph{No repeated stochastic trials.}
Each query is executed once at temperature $0$. Hosted inference can still
exhibit residual nondeterminism, so the present experiment does not estimate
run-to-run variance.

\paragraph{No component-level ablation.}
The experiment compares the complete ReAdapt protocol against ReAct. It does
not separately identify the contribution of the relationship state, explicit
state deltas, policy operations, or the additional structured reasoning step.

\paragraph{Compute is not perfectly matched.}
ReAdapt generates an additional Adapt block after each observation and uses
slightly more tool calls on average. The experiment therefore isolates a
reasoning-protocol intervention under matched model and information access,
not equal token-level computation.

\paragraph{Not all state dimensions are independently exercised.}
The current tasks primarily test belief and relationship state.
Goal, norm, and disclosure are included to support broader social-agent
settings, but their independent value remains to be evaluated on tasks that
explicitly require them.

\section{Discussion}
\label{sec:discussion}

The results point to an important distinction between
\emph{having access to relational information} and
\emph{allowing relational information to change a decision}. Both ReAct and
ReAdapt can retrieve friendship, follow, reciprocity, and interaction signals
from the same underlying social environment. Yet making those observations
available in context does not guarantee that they will revise an already
plausible candidate.

The overturn diagnostic makes this distinction particularly visible. When
surface and relational signals agree, an agent can succeed without revealing
whether it modeled the relationship at all. When they disagree, success
requires hypothesis revision. ReAdapt provides an explicit interface for that
revision: observations update state, state changes can emit
\textsc{switch}, and the revised state conditions the subsequent action.

This perspective also suggests that relationship reasoning should not be
viewed solely as an information-retrieval problem. Providing an agent with
access to a social graph is useful only if retrieved evidence is coupled
reliably to the decision process. ReAdapt treats tool observations as updates
to an evolving social state rather than as additional unstructured context.

More broadly, the framework points toward personal agents that maintain
persistent models of goals, relationships, norms, and disclosure boundaries.
The present benchmark studies only a controlled subset of this broader
problem. Real deployment would additionally require calibrated uncertainty,
privacy safeguards, temporal adaptation, user correction, and evaluation
against human behavioral outcomes.

\end{document}